\documentclass[11pt,a4paper]{article}
\usepackage{times,latexsym}
\usepackage{url}
\usepackage[T1]{fontenc}

\usepackage{graphicx}
\usepackage{amsmath}
\usepackage{amsfonts}
\usepackage{amssymb}
\usepackage{bbm}
\usepackage{amsthm}
\usepackage{algorithmicx}
\usepackage{algorithm, algpseudocode}
\usepackage{booktabs}
\usepackage{enumitem}
\usepackage{multirow}
\usepackage{tikz}
\usepackage[most]{tcolorbox}
\usepackage{algorithm, algpseudocode}
\usepackage{bbm}
\usepackage[figuresright]{rotating}
\usepackage{longtable}
\usepackage[table]{xcolor}
\usepackage{colortbl}
\usepackage{adjustbox}
\definecolor{darkgreen}{RGB}{34, 139, 34}   % Forest green - slightly brighter
\definecolor{darkred}{RGB}{205, 92, 92}

\newtheorem{definition}{Definition}[section]

\usepackage[acceptedWithA]{tacl2021v1}
\usepackage{xspace,mfirstuc,tabulary}

\newif\iftaclinstructions
\taclinstructionsfalse % AUTHORS: do NOT set this to true
\iftaclinstructions
\renewcommand{\confidential}{}
\renewcommand{\anonsubtext}{(No author info supplied here, for consistency with
TACL-submission anonymization requirements)}
\newcommand{\instr}
\fi

\iftaclpubformat % this "if" is set by the choice of options

\else

\fi

\title{Bridging Auxiliary Constraints to Resolve Instruction Following\\in Large Reasoning Models}

\author{
  \textbf{Zhengyi Zhao\textsuperscript{1}},
  \textbf{Shubo Zhang\textsuperscript{1}},
  \textbf{Huimin Wang\textsuperscript{3}},
  \textbf{Zezhong Wang\textsuperscript{1}},
  \textbf{Yutian Zhao\textsuperscript{3}},\\
  \textbf{Yefeng Zheng\textsuperscript{4}},
  \textbf{Binyang Li\textsuperscript{2}},
  \textbf{Yulan He\textsuperscript{5}},
  \textbf{Kam-Fai Wong\textsuperscript{1}},
  \textbf{Xian Wu\textsuperscript{3}}
\\
  \textsuperscript{1} The Chinese University of Hong Kong
  \textsuperscript{2} University of International Relations \\
  \textsuperscript{3} Tencent Jarvis Lab
  \textsuperscript{4} Westlake University
  \textsuperscript{5} King's College London
\\
  {
  \texttt{zyzhao@se.cuhk.edu.hk}
  }
}

\date{}

\begin{document}
\maketitle
\begin{abstract}
Large Reasoning Models (LRMs) have demonstrated impressive capabilities in many tasks, yet they struggle with reliably following multiple instructions, either by failing to satisfy individual constraints or by struggling to balance competing constraints simultaneously. We formalize this challenge as the Constraint Adherence Problem (CAP). This paper introduces a novel framework that addresses CAP by representing instructions as a structured knowledge graph of constraints. Our approach, Constraint Relationship Graph Completion (CRGC), explicitly models relationships between constraints, identifies adherence challenges, and discovers ``bridge constraints'' that help the model better focus on and reconcile requirements. Bridge constraints act as auxiliary instructions that make primary constraints more salient and compatible. Unlike existing approaches that enhance instruction following through general training methods, CRGC specifically improves constraint satisfaction by leveraging the model's own knowledge to create better pathways for generation. Experiments across three popular instruction following datasets demonstrate that our approach reduces constraint violations by 39\% compared to standard prompting while maintaining reasoning abilities of large reasoning models.
\end{abstract}

\section{Introduction}

Large Reasoning Models (LRMs) have shown remarkable abilities in responding to instructions, yet they frequently struggle with fully satisfying all requirements in a given prompt~\cite{honovich-etal-2023-instruction,peng-etal-2024-answer,wei-etal-2025-ji2s,luo-etal-2025-tree}. This challenge shows in two key ways: (1) models sometimes overlook certain constraints entirely~\cite{zhang-etal-2024-star,shi-etal-2024-autodsl,pei-etal-2025-mathfusion}, and (2) models often fail to balance multiple constraints that appear to conflict~\cite{pan-etal-2023-fact,jiang-etal-2024-followbench,li-etal-2025-open}. For example, when asked to ``write a detailed technical explanation about quantum computing within 200 words,'' an LRM might produce text that is either too detailed (violating word limit) or too simplified (violating technical detail). In other cases, a model might completely overlook a constraint such as ``enclose your answer with double quotation marks'' even when there's no apparent conflict with other requirements.

We formalize this widespread challenge as the \textit{Constraint Adherence Problem (CAP)}, which represents a significant barrier to reliable deployment of LRMs in real-world applications~\cite{ouyang2022training,zhao-etal-2025-memereacon}. While researchers have developed various approaches to improve instruction following through supervised fine-tuning~\cite{wu-etal-2023-learning,du-chilton-2023-storywars}, reinforcement learning from human feedback~\cite{shi-etal-2024-opex,liu-etal-2025-rag}, and prompt engineering techniques~\cite{yin-etal-2023-read,srikanth-etal-2024-often,held-etal-2025-distilling}, these methods don't specifically address how models recognize, prioritize, and reconcile multiple constraints. Instead, they implicitly assume that better general capabilities will naturally lead to better constraint satisfaction, an assumption that may not always hold in practice.

Our key insight is that effective constraint satisfaction requires explicitly modeling how constraints relate to each other and to the overall task. To this end, we introduce a structured approach where each instruction is broken down into specific constraints that form a relationship graph. In this graph, nodes represent individual constraints, and edges represent how constraints interact with each other, supporting, interfering with, or acting independently of one another.

This graph-based representation reveals that LRMs struggle most when constraints are either insufficiently salient or perceived as contradictory, termed as constraint interference. Based on this insight, we develop the \textit{Constraint Relationship Graph Completion} (CRGC) framework with three main components: (1) \textbf{Constraint Graph Construction} that maps relationships between decomposed instruction constraints; (2) \textbf{Adherence Challenge Detection} that identifies overlooked or conflicting constraints; and (3) \textbf{Bridge Constraint Discovery} that introduces auxiliary instructions to connect or reconcile problematic constraints. We define \textit{bridge constraints} as intermediate instructions that address conceptual gaps causing model failures by creating logical connections between conflicting or disconnected primary constraints. Importantly, we express bridge constraints in natural language rather than categorizing them into predefined types, preventing artificial restrictions on model behavior and avoiding the deployment of multiple bridges for a single constraint pair. For example, when a model struggles to write text that is both ``comprehensive'' and ``within 200 words,'' a bridge constraint like ``use structured sections with bullet points for key details'' provides a concrete strategy to satisfy both requirements. Our approach leverages the model's knowledge to enhance constraint satisfaction without requiring additional training or complex prompting strategies.

We evaluated our approach across instruction following benchmarks. Experimental results demonstrate that CRGC reduces constraint violations by 39\% compared to standard prompting and by 27.7\% compared to Chain-of-Thought (CoT) prompting, while maintaining or improving output quality as assessed by human evaluators. Notably, unlike existing instruction-following methods that typically sacrifice performance on general reasoning tasks after specialization, our approach shows no degradation on standard reasoning benchmarks. This advantage stems from CRGC's non-invasive nature, it enhances constraint satisfaction without modifying model parameters and adaptively determines when bridge constraints are beneficial, automatically omitting them when unnecessary. The primary contributions of this paper are:
\begin{itemize}[itemsep=0pt,parsep=0pt,topsep=0pt,partopsep=0pt]
    \item A formal characterization of the Constraint Adherence Problem in LRMs, primarily focusing on constraint oversight and constraint interference problems
    \item A graph-based framework for representing constraint relationships and identifying adherence challenges, along with an effective method for discovering bridge constraints.
    \item Empirical evidence demonstrating great improvements in constraint satisfaction across diverse datasets without requiring model retraining, while maintaining the reasoning abilities.
\end{itemize}

\section{Related Works}

Recent research has focused on evaluating and enhancing instruction following capabilities in LLMs. Evaluation benchmarks such as \citet{jiang-etal-2024-followbench} and \citet{qin2024infobench} assess hierarchical and complex instruction following, while \citet{wu-etal-2025-lifbench} proposed metrics for cross-context consistency. Parameter-update approaches have improved instruction adherence through techniques like reinforcement learning from human feedback \citep{ouyang2022training} and instruction tuning \citep{wei2022finetuned,chung2024scaling,longpre2023flan}. However, these methods typically treat instruction following as a monolithic skill without explicitly modeling relationships between potentially conflicting constraints.

Multi-attribute steering represents another approach to control specific aspects of LLM outputs. \citet{Dathathri2020Plug} and \citet{krause-etal-2021-gedi-generative} introduced plug-and-play frameworks for controlling attributes like toxicity and sentiment, while \citet{subramani-etal-2022-extracting} identified latent directions in embedding space corresponding to specific attributes. More recent work by \citet{wang2024controllable} developed techniques for simultaneously controlling multiple attributes through vector composition. While these approaches share our goal of satisfying multiple requirements, our work differs in several key aspects: (1) we address arbitrary constraints with complex interdependencies rather than predefined independent attributes, (2) we leverage the model's reasoning capabilities through bridge constraints instead of modifying internal representations, and (3) we explicitly model constraint relationships to identify which ones need reconciliation, making our approach more adaptive to diverse instruction types where constraints may be implicit, conflicting, or hierarchical.

\begin{figure*}[!t]
    \includegraphics[width=\linewidth]{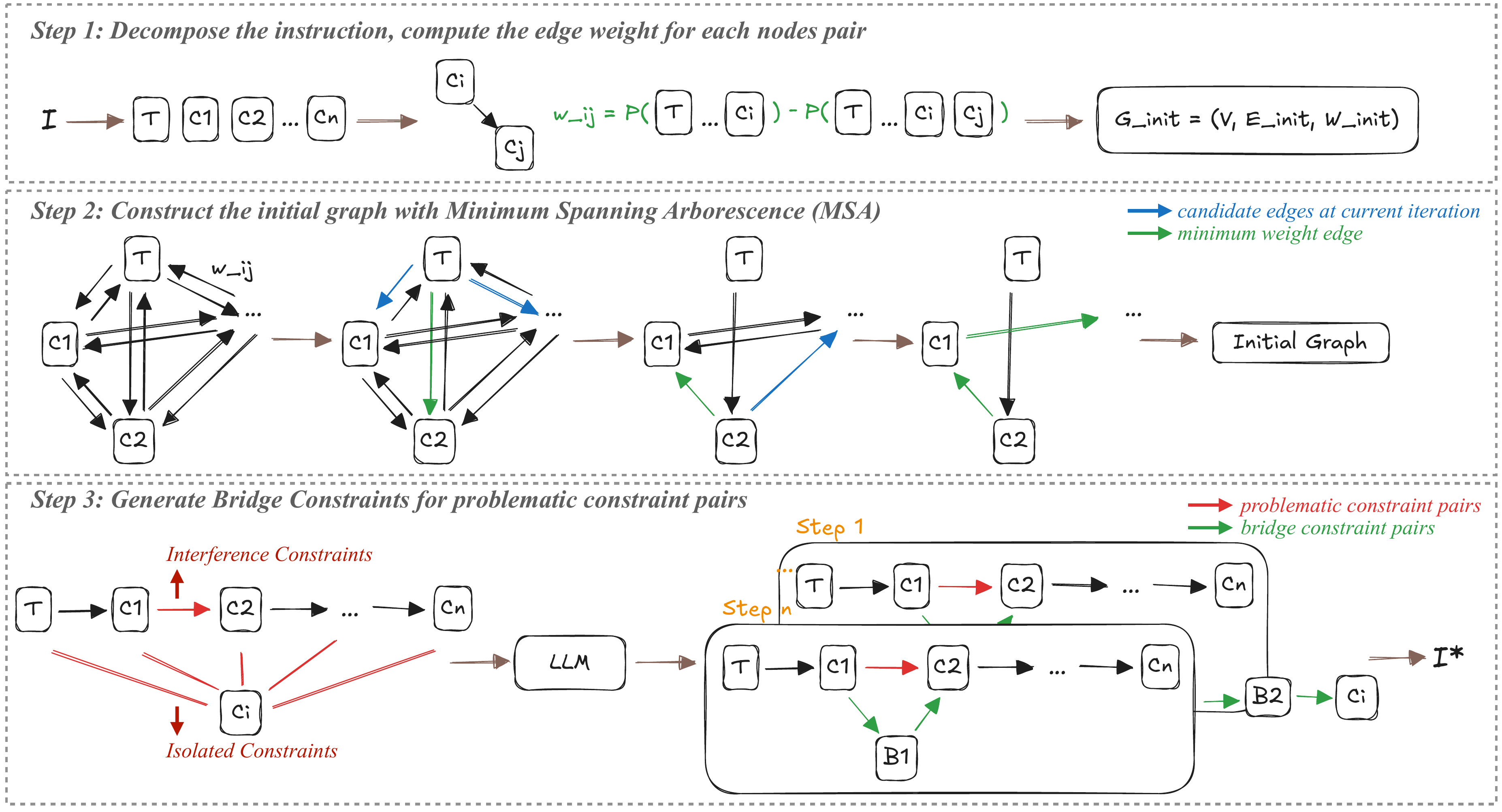}
    \caption{Overview of the Constraint Relationship Graph Completion (CRGC) framework. Step 1: The input instruction $I$ is decomposed into individual constraints $\{C_1, C_2, \ldots, C_n\}$, and edge weights $w_{ij}$ are computed based on the leave-one-out probabilities to form the initial graph $G_{init}$. Step 2: The Minimum Spanning Arborescence (MSA) algorithm~\cite{edmonds1967optimum} constructs an optimized constraint graph by iteratively selecting minimum weight edges (green) from candidate edges (blue) to connect all constraint nodes to the task node $T$. Step 3: Problematic constraint pairs (with red arrows) are identified, including interference constraints and isolated constraints. Bridge constraints $B_1, B_2$ are then generated using an LLM to address these problematic pairs, resulting in an enhanced instruction $I^*$ that improves constraint adherence.}
    \label{fig:overview}
\end{figure*}

A parallel line of research focuses on post-hoc iterative refinement to improve constraint adherence. Frameworks such as DeCRIM \citep{ferraz2024llm} and Divide-Verify-Refine \citep{zhang2025divide} employ a multi-stage pipeline where an initial draft is generated, evaluated against constraints, and subsequently refined to correct identified violations. While these iterative correction methods improve upon zero-shot prompting, they operate reactively. Because they lack an explicit, holistic model of how constraints interact, fixing one violation often inadvertently triggers a new violation in a previously satisfied constraint. In contrast, our CRGC framework operates proactively. By constructing a comprehensive relationship graph prior to generation, CRGC identifies inherent conflicts and deploys bridge constraints to structurally reconcile them during the initial forward pass. This graph-based foresight not only yields higher ultimate satisfaction rates for conflicting constraints but fundamentally bypasses the computationally expensive reactive correction.

\section{Methodology}
In this section, we present our Constraint Relationship Graph Completion (CRGC) framework in detail. We first formalize the Constraint Adherence Problem (CAP) and introduce the concept of directed constraint relationships. Then, we describe our three-component approach: (1) Constraint Graph Construction, (2) Adherence Challenge Detection, and (3) Bridge Constraint Discovery.

\subsection{Problem Formulation}
Given a language model $M$ that maps instructions $I$ to output text $y \in \mathcal{Y}$, we define a constraint $C$ as a specific requirement that the output text must satisfy. An instruction $I$ typically contains a core task $\mathbf{R}$ and multiple constraints $\mathcal{C} = \{C_1, C_2, \dots, C_n\}$. The Constraint Adherence Problem (CAP) occurs when a model fails to satisfy all constraints in an instruction, either by overlooking certain constraints entirely or by failing to balance multiple constraints that appear to conflict.

\subsubsection{Constraint Satisfaction}
For a given constraint $C_i$ and output text $y$, we define a satisfaction function $S(C_i, y) \in [0, 1]$ by prompting an evaluation Large Language Model (LLM) as the judge, where 1 indicates perfect satisfaction and 0 indicates complete violation. For binary constraints (e.g., ``enclose answer in quotes''), $S(C_i, y)$ is either 0 or 1. For continuous constraints (e.g., ``make explanation detailed''), $S(C_i, y)$ can take intermediate values. The overall constraint satisfaction for an instruction is given by the bottleneck performance:
\begin{equation}
S_{\rm total}(I, y) = \min_{C_i \in \mathcal{C}} S(C_i, y).
\end{equation}
We use the minimum rather than the average because satisfying most constraints while completely violating one remains a failure in strict instruction following.

\subsubsection{Directed Constraint Influences}
Constraint relationships in autoregressive generation are inherently asymmetric. We characterize three fundamental directed relationships based on how satisfying constraint $C_i$ influences the expected satisfaction of constraint $C_j$. Let $\mathbb{E}[S(C_j, y)]$ denote the baseline expected satisfaction, and $\mathbb{E}[S(C_j, y) \mid S(C_i, y)\ge \tau]$\footnote{Here, we set $\tau$ to strictly satisfy the $i$-th constraint.} denote the conditional expectation given that $C_i$ is highly satisfied, where $\tau$ denotes the relaxation rate to evaluate whether or not the instruction can be regarded as followed.

\begin{definition}[Enhancing Influence]
Constraint $C_i$ enhances $C_j$ if satisfying $C_i$ strictly increases the the expected satisfaction of $C_j$:
\begin{equation}
\mathbb{E}[S(C_j, y) \mid S(C_i, y) \ge \tau] > \mathbb{E}[S(C_j, y)].
\end{equation}
\end{definition}

\begin{definition}[Interfering Influence]
Constraint $C_i$ interferes with $C_j$ if satisfying $C_i$ decreases the expected satisfaction of $C_j$, representing a perceived conflict:
\begin{equation}
\mathbb{E}[S(C_j, y) \mid S(C_i, y) \ge \tau] < \mathbb{E}[S(C_j, y)].
\end{equation}
\end{definition}

\begin{definition}[Neutral / Independent Influence]
Constraint $C_i$ has a neutral influence on $C_j$ if satisfying $C_i$ does not meaningfully affect the satisfaction of $C_j$:
\begin{equation}
|\mathbb{E}[S(C_j, y) \mid S(C_i, y) \ge \tau] - \mathbb{E}[S(C_j, y)]| \leq \varepsilon,
\end{equation}
where $\varepsilon$ is a small margin of independence. Note that this is a directed property of the edge $C_i \to C_j$, not a symmetric global independence.
\end{definition}

\subsection{Constraint Relationship Graph Completion Framework}

Our CRGC framework systematically addresses the CAP through three sequential steps: constructing a constraint relationship graph, detecting adherence challenges, and discovering appropriate bridge constraints. This framework leverages the model's own capabilities to enhance constraint satisfaction without requiring parameter updates.

\subsubsection{Constraint Graph Construction}
The first step decomposes instructions into individual constraints to construct a weighted directed graph representing their influences. Given an instruction $I$, we construct an initial fully connected directed graph $G_{\text{init}} = (V, E_{\text{init}}, W_{\text{init}})$, where the vertex set $V = \{C_1, C_2, \dots, C_n\} \cup \{\mathbf{T}\}$ comprises all constraints plus the overall task objective. The edge set $E_{\text{init}} = \{(C_i, C_j) \mid C_i, C_j \in V, i \neq j\}$ contains directed connections.

To maximize overall constraint satisfaction, we want to find a generation pathway that maximizes enhancing relationships and avoids interfering ones. Thus, we define the edge weight $w_{ij} \in W_{\text{init}}$ as the \textit{cost} (or risk of violation) of attempting to satisfy $C_j$ after $C_i$. This is quantified by the relative drop in expected satisfaction:
\begin{equation}
w_{ij} = \frac{\mathbb{E}[S(C_j, y)] - \mathbb{E}[S(C_j, y) \mid S(C_i, y) \ge \tau]}{\mathbb{E}[S(C_j, y)]}.
\end{equation}
% Here, $\alpha$ is a small Laplace smoothing factor (e.g., $10^{-3}$) added to prevent division by zero and stabilize the estimator.
To ensure statistically meaningful estimates, we generate $N$ candidate outputs using a non-zero temperature to capture diverse generation paths\footnote{Here we set $N$ equals to 11.}. 

Under this formulation, if $C_i$ enhances $C_j$, the cost $w_{ij}$ is negative (beneficial). If $C_i$ interferes with $C_j$, the cost $w_{ij}$ is positive (detrimental). We then apply the Minimum Spanning Arborescence (MSA) algorithm~\cite{edmonds1967optimum} to find a directed spanning tree rooted at $\mathbf{R}$ with minimum total cost. This directly aligns the MSA objective with our goal: producing a directed tree that minimizes constraint violation risk. Formally, we seek the tree $T^* = (V, E_T, W_T)$ such that:
\begin{equation}
T^* = \arg\min_{T' \in T} \sum_{(i,j) \in E_{T'}} w_{ij},
\end{equation}
where $T$ is the set of all possible directed spanning trees rooted at $\mathbf{R}$. 

\subsubsection{Adherence Challenge Detection}

The second step analyzes the optimized tree $T^*$ to identify constraint relationships that still present challenges. We categorize edges in the tree based on their cost weights:
\begin{equation}
\text{Category}(i,j) = 
\begin{cases}
\text{Enhance}, & \text{if } w_{ij} < -\delta, \\
\text{Independent}, & \text{if } |w_{ij}| \leq \delta, \\
\text{Interfere}, & \text{if } w_{ij} > \delta,
\end{cases}
\end{equation}
where $\delta$ is an empirically determined positive threshold\footnote{Here we set $\delta$ equals to 0.3 as default.}. Interfering edges represent constraints that the model perceives as contradictory despite the optimized path, while independent edges represent disjoint constraints that the model fails to cognitively connect, leading to overlooked requirements.

We identify the set of problematic edges $E_{\text{prob}} \subset E_T$ as those categorized as interfering or independent. These edges become candidates for bridge constraint insertion.

\subsubsection{Bridge Constraint Discovery}

The final step addresses the identified adherence challenges by generating intermediate natural language instructions. For each problematic edge $(C_i, C_j) \in E_{\text{prob}}$, we generate a bridge constraint $B_{ij}$ using the language model $M$:
\begin{equation}
B_{ij} = \begin{cases}
\mathcal{R}(C_i, C_j), & \text{if } w_{ij} > \delta \text{ (interfering)} \\
\mathcal{C}(C_i, C_j), & \text{if } |w_{ij}| \leq \delta \text{ (independent)},
\end{cases}
\end{equation}
where $\mathcal{R}(\cdot,\cdot)$ and $\mathcal{C}(\cdot,\cdot)$ are reconciliation and connection prompt templates, respectively.

For interfering constraints, $\mathcal{R}(\cdot,\cdot)$ guides the model to resolve perceived conflicts. For independent constraints, $\mathcal{C}(\cdot,\cdot)$ establishes explicit relationships. For example, given interfering constraints ``be concise'' ($C_i$) and ``provide detailed explanations'' ($C_j$), the model might generate the bridge ``focus on key information with precise language while ensuring all necessary details are included.''

We incorporate these bridges to create an enhanced structure $G_{\text{enhanced}} = (V_{\text{enhanced}}, E_{\text{enhanced}}, W_{\text{enhanced}})$, where $V_{\text{enhanced}} = V \cup \{B_{ij} \mid (C_i, C_j) \in E_{\text{prob}}\}$. Problematic edges are replaced with bridge-mediated connections: $E_{\text{enhanced}} = (E_T \setminus E_{\text{prob}}) \cup \{(C_i, B_{ij}), (B_{ij}, C_j) \mid (C_i, C_j) \in E_{\text{prob}}\}$. 

Finally, we augment the original instruction by seamlessly context-integrating the bridge constraints:
\begin{equation}
I_{\text{enhanced}} = I \oplus \{B_{ij} \mid (C_i, C_j) \in E_{\text{prob}}\},
\end{equation}
where $\oplus$ denotes the concatenation and formatting of the new constraints into the prompt payload. This targeted intervention resolves adherence failures dynamically, allocating computational overhead strictly to the nodes that exhibit conceptual friction.

\section{Experiments}

\subsection{Experimental Settings}

\paragraph{Datasets.} We evaluate CRGC on three instruction following benchmarks and three general reasoning benchmarks to demonstrate both improved constraint adherence and preserved reasoning capabilities.

Our evaluation employs three comprehensive instruction following benchmarks. IFEval~\cite{zhou2023instructionfollowingevaluationlargelanguage} provides a comprehensive benchmark with 541 instructions across 25 categories, designed to evaluate various aspects of instruction following abilities. ComplexBench~\cite{NEURIPS2024_f8c24b08} contains 1,150 complex, multi-constraint instructions that require models to balance competing requirements. FollowBench~\cite{jiang-etal-2024-followbench} features 820 instruction-response pairs with explicit constraints and evaluation criteria focusing on instruction handling capabilities.

To ensure our method preserves reasoning abilities, we utilize three standard reasoning benchmarks. MMLU~\cite{hendrycks2021measuring} serves as a multitask benchmark covering 57 subjects across STEM, humanities, social sciences, and more. GSM8K~\cite{cobbe2021trainingverifierssolvemath} contains 8,500 high-quality grade school math problems requiring multi-step reasoning. BIG-Bench Hard (BBH)~\cite{suzgun-etal-2023-challenging}, consists of 23 challenging tasks selected from BIG-Bench focusing on complex reasoning.

\paragraph{Evaluation Metrics.} For comprehensive evaluation, we employ several complementary metrics across different dimensions. To ensure rigorous assessment and prevent evaluation circularity, our evaluation pipeline processes constraints according to their type. First, for strictly quantitative and structural constraints (e.g., word count limits, specific formatting, presence of keywords), we utilize deterministic Python scripts (e.g., exact token counters and regex) to strictly bypass the known limitations of LLMs on exact counting tasks. Second, for semantic constraints requiring qualitative judgment, we employ a decoupled evaluation framework. The evaluation model is strictly isolated from the generation model (e.g., using Claude-3.7-Sonnet to evaluate GPT-4o outputs, and vice versa) to eliminate self-preference bias.

We evaluate constraint adherence using two primary metrics. Constraint Satisfaction Rate (CSR) measures the proportion of constraints fully satisfied in a response:
\begin{equation}
    \text{CSR} = \frac{1}{|I|} \sum_{i=1}^{|I|} \frac{1}{|\mathcal{C}_i|} \sum_{j=1}^{|\mathcal{C}_i|} \mathbbm{1}[S(C_{ij}, y_i) \geq \tau],
\end{equation}
where $I$ is the set of all instructions, $\mathcal{C}_i$ is the set of constraints for instruction $i$, $S(C_{ij}, y_i)$ is the satisfaction score assigned by our evaluation pipeline for constraint $j$ of instruction $i$, and $\tau$ is the satisfaction threshold. Weighted Constraint Satisfaction (WCS) provides a more nuanced evaluation by accounting for partial satisfaction of continuous constraints:
\begin{equation}
    \text{WCS} = \frac{1}{|I|} \sum_{i=1}^{|I|} \frac{1}{|\mathcal{C}_i|} \sum_{j=1}^{|\mathcal{C}_i|} S(C_{ij}, y_i),
\end{equation}
where $S(C_{ij}, y_i) \in [0,1]$ represents the degree of satisfaction.

To assess overall task effectiveness, we use multiple performance indicators. Task Completion Rate (TCR) measures the proportion of tasks completed successfully:
\begin{equation}
    \text{TCR} = \frac{1}{|I|} \sum_{i=1}^{|I|}S_{\text{total}}(I_i, y_i),
\end{equation}
where $S_{\text{total}}(I_i, y_i)$ is the overall satisfaction score for instruction $i$. Standard accuracy metrics are applied for reasoning tasks, while F1 Score is used for tasks with precision-recall tradeoffs.

\begin{table*}[!t]
\centering
\caption{Performance comparison on three instruction following benchmarks. Best performance for each model is highlighted in \textbf{bold}. CSR, WCS, TCR stand for constraint-level, weighted-constraint-level, and task-level performance, respectively. The baselines SP, CoT, SR denotes Standard Prompting, Chain-of-Thoughts, and Self-Reflection, respectively.}
\label{tab:main_results}
\begin{adjustbox}{width=\textwidth}
\begin{tabular}{lccccccccc}
\toprule
\multirow{2}{*}{\textbf{Method}} & \multicolumn{3}{c}{\textbf{IFEval}} & \multicolumn{3}{c}{\textbf{ComplexBench}} & \multicolumn{3}{c}{\textbf{FollowBench}} \\
\cmidrule(lr){2-4} \cmidrule(lr){5-7} \cmidrule(lr){8-10}
& \textbf{CSR} & \textbf{WCS} & \textbf{TCR} & \textbf{CSR} & \textbf{WCS} & \textbf{TCR} & \textbf{CSR} & \textbf{WCS} & \textbf{TCR} \\
\midrule
\multicolumn{10}{c}{\textit{Closed-source Models}} \\
\midrule
GPT-4o + SP & 78.9$\pm$1.2 & 82.5$\pm$1.0 & 80.3$\pm$1.1 & 66.7$\pm$1.5 & 70.8$\pm$1.3 & 68.1$\pm$1.4 & 73.2$\pm$1.3 & 77.3$\pm$1.1 & 74.5$\pm$1.2 \\
GPT-4o + CoT & 82.7$\pm$0.9 & 86.1$\pm$0.8 & 84.2$\pm$0.9 & 71.5$\pm$1.2 & 75.4$\pm$1.0 & 72.8$\pm$1.1 & 77.1$\pm$1.0 & 80.9$\pm$0.9 & 78.3$\pm$1.0 \\
GPT-4o + SR & 84.5$\pm$0.8 & 87.8$\pm$0.7 & 86.1$\pm$0.8 & 73.2$\pm$1.0 & 77.1$\pm$0.9 & 74.6$\pm$1.0 & 78.9$\pm$0.9 & 82.7$\pm$0.8 & 80.1$\pm$0.9 \\
GPT-4o + CRGC & \textbf{88.7$\pm$0.6} & \textbf{91.3$\pm$0.5} & \textbf{89.6$\pm$0.7} & \textbf{78.6$\pm$0.8} & \textbf{82.5$\pm$0.7} & \textbf{79.8$\pm$0.9} & \textbf{83.8$\pm$0.7} & \textbf{87.4$\pm$0.6} & \textbf{84.9$\pm$0.8} \\
\midrule
Claude-3.7-Sonnet + SP & 79.5$\pm$1.2 & 83.1$\pm$0.9 & 81.2$\pm$1.1 & 67.4$\pm$1.4 & 71.6$\pm$1.2 & 68.8$\pm$1.3 & 74.3$\pm$1.3 & 78.2$\pm$1.1 & 75.7$\pm$1.2 \\
Claude-3.7-Sonnet + CoT & 83.6$\pm$0.9 & 87.0$\pm$0.7 & 85.2$\pm$0.8 & 72.3$\pm$1.1 & 76.2$\pm$0.9 & 73.7$\pm$1.0 & 78.4$\pm$1.0 & 82.1$\pm$0.8 & 79.5$\pm$0.9 \\
Claude-3.7-Sonnet + SR & 85.7$\pm$0.7 & 88.9$\pm$0.6 & 87.3$\pm$0.7 & 74.1$\pm$0.9 & 78.0$\pm$0.8 & 75.5$\pm$0.9 & 80.2$\pm$0.8 & 84.0$\pm$0.7 & 81.4$\pm$0.8 \\
Claude-3.7-Sonnet + CRGC & \textbf{90.2$\pm$0.5} & \textbf{92.7$\pm$0.4} & \textbf{91.0$\pm$0.6} & \textbf{79.8$\pm$0.7} & \textbf{83.6$\pm$0.6} & \textbf{81.0$\pm$0.8} & \textbf{85.1$\pm$0.6} & \textbf{88.6$\pm$0.5} & \textbf{86.3$\pm$0.7} \\
\midrule
Gemini-2.5-Pro + SP & 78.2$\pm$1.3 & 81.9$\pm$1.0 & 79.8$\pm$1.2 & 65.8$\pm$1.5 & 70.0$\pm$1.3 & 67.2$\pm$1.4 & 72.7$\pm$1.4 & 76.8$\pm$1.2 & 74.0$\pm$1.3 \\
Gemini-2.5-Pro + CoT & 82.0$\pm$1.0 & 85.5$\pm$0.8 & 83.6$\pm$0.9 & 70.6$\pm$1.2 & 74.5$\pm$1.0 & 72.0$\pm$1.1 & 76.5$\pm$1.1 & 80.3$\pm$0.9 & 77.8$\pm$1.0 \\
Gemini-2.5-Pro + SR & 83.8$\pm$0.8 & 87.1$\pm$0.7 & 85.5$\pm$0.8 & 72.3$\pm$1.0 & 76.3$\pm$0.9 & 73.8$\pm$1.0 & 78.2$\pm$0.9 & 82.0$\pm$0.8 & 79.5$\pm$0.9 \\
Gemini-2.5-Pro + CRGC & \textbf{88.0$\pm$0.6} & \textbf{90.7$\pm$0.5} & \textbf{89.1$\pm$0.7} & \textbf{77.5$\pm$0.8} & \textbf{81.4$\pm$0.7} & \textbf{78.9$\pm$0.9} & \textbf{83.0$\pm$0.7} & \textbf{86.7$\pm$0.6} & \textbf{84.2$\pm$0.8} \\
\midrule
\multicolumn{10}{c}{\textit{Open-source Models}} \\
\midrule
Qwen2.5-72B + SP & 74.3$\pm$1.5 & 78.0$\pm$1.3 & 75.8$\pm$1.4 & 62.5$\pm$1.7 & 66.9$\pm$1.5 & 63.9$\pm$1.6 & 69.2$\pm$1.6 & 73.5$\pm$1.4 & 70.6$\pm$1.5 \\
Qwen2.5-72B + CoT & 78.1$\pm$1.2 & 81.7$\pm$1.0 & 79.4$\pm$1.1 & 66.8$\pm$1.4 & 70.9$\pm$1.2 & 68.2$\pm$1.3 & 72.9$\pm$1.3 & 77.1$\pm$1.1 & 74.2$\pm$1.2 \\
Qwen2.5-72B + SR & 80.3$\pm$1.0 & 83.7$\pm$0.9 & 81.5$\pm$1.0 & 68.9$\pm$1.2 & 73.0$\pm$1.0 & 70.3$\pm$1.1 & 74.8$\pm$1.1 & 78.9$\pm$0.9 & 76.0$\pm$1.0 \\
Qwen2.5-72B + CRGC & \textbf{85.5$\pm$0.8} & \textbf{88.9$\pm$0.7} & \textbf{86.7$\pm$0.9} & \textbf{74.8$\pm$1.0} & \textbf{78.7$\pm$0.8} & \textbf{76.1$\pm$1.0} & \textbf{80.4$\pm$0.9} & \textbf{84.2$\pm$0.7} & \textbf{81.5$\pm$0.9} \\
\midrule
Llama-3.1-72B + SP & 75.1$\pm$1.4 & 78.8$\pm$1.2 & 76.5$\pm$1.3 & 63.4$\pm$1.6 & 67.7$\pm$1.4 & 64.7$\pm$1.5 & 70.0$\pm$1.5 & 74.2$\pm$1.3 & 71.4$\pm$1.4 \\
Llama-3.1-72B + CoT & 79.0$\pm$1.1 & 82.5$\pm$0.9 & 80.3$\pm$1.0 & 67.8$\pm$1.3 & 71.9$\pm$1.1 & 69.1$\pm$1.2 & 73.8$\pm$1.2 & 77.9$\pm$1.0 & 75.0$\pm$1.1 \\
Llama-3.1-72B + SR & 81.2$\pm$0.9 & 84.5$\pm$0.8 & 82.4$\pm$0.9 & 69.7$\pm$1.1 & 73.8$\pm$0.9 & 71.0$\pm$1.0 & 75.6$\pm$1.0 & 79.6$\pm$0.8 & 76.8$\pm$0.9 \\
Llama-3.1-72B + CRGC & \textbf{86.3$\pm$0.7} & \textbf{89.5$\pm$0.6} & \textbf{87.4$\pm$0.8} & \textbf{75.6$\pm$0.9} & \textbf{79.4$\pm$0.7} & \textbf{76.8$\pm$0.9} & \textbf{81.2$\pm$0.8} & \textbf{84.9$\pm$0.6} & \textbf{82.3$\pm$0.8} \\
\midrule
Mixtral-8x22B + SP & 72.4$\pm$1.6 & 76.1$\pm$1.4 & 73.9$\pm$1.5 & 60.3$\pm$1.8 & 64.5$\pm$1.6 & 61.7$\pm$1.7 & 67.2$\pm$1.7 & 71.4$\pm$1.5 & 68.5$\pm$1.6 \\
Mixtral-8x22B + CoT & 76.0$\pm$1.3 & 79.6$\pm$1.1 & 77.3$\pm$1.2 & 64.5$\pm$1.5 & 68.6$\pm$1.3 & 65.8$\pm$1.4 & 70.8$\pm$1.4 & 74.9$\pm$1.2 & 72.1$\pm$1.3 \\
Mixtral-8x22B + SR & 78.1$\pm$1.1 & 81.5$\pm$0.9 & 79.3$\pm$1.0 & 66.4$\pm$1.3 & 70.5$\pm$1.1 & 67.7$\pm$1.2 & 72.5$\pm$1.2 & 76.6$\pm$1.0 & 73.8$\pm$1.1 \\
Mixtral-8x22B + CRGC & \textbf{83.4$\pm$0.9} & \textbf{86.7$\pm$0.8} & \textbf{84.5$\pm$1.0} & \textbf{72.2$\pm$1.1} & \textbf{76.1$\pm$0.9} & \textbf{73.5$\pm$1.1} & \textbf{78.3$\pm$1.0} & \textbf{82.0$\pm$0.8} & \textbf{79.4$\pm$0.9} \\
\bottomrule
\end{tabular}
\end{adjustbox}
\end{table*}

\paragraph{Baselines.} We compare CRGC against state-of-the-art models and instruction following methods.

Our evaluation includes both open-source and closed-source models. For open-source models, we utilize Qwen2.5-72B~\cite{qwen2025qwen25technicalreport}, Llama-3.1-70B~\cite{llama31}, and Mixtral-8x22B~\cite{jiang2024mixtralexperts}. The closed-source models in our comparison include GPT-4o~\cite{hurst2024gpt}, Claude-3.7-Sonnet-Thinking~\cite{Claude3S}, Gemini-2.5-Pro~\cite{comanici2025gemini}.

We compare against several established instruction following techniques. Standard Prompting (SP) serves as our baseline, using direct instruction input without additional techniques~\cite{wei2022chain}. CoT prompts models to show step-by-step reasoning~\cite{wei2022chain}. Self-Consistency (SC) generates multiple responses and selects the most consistent output~\cite{wang2023selfconsistency}. Self-Reflection (SR) prompts models to reflect on their outputs against requirements~\cite{ji-etal-2023-towards}. Constraint Optimization Prompting (COP) implements explicit optimization for constraint satisfaction~\cite{long-etal-2024-prompt}. Think-to-Think (T$^2$) solve the problem by reflecting multiple samples to model for reference~\cite{zhao-etal-2025-t2}.

\paragraph{Implementation Details.} We implemented automated constraint extraction using GPT-4o decomposition. We computed the Minimum Spanning Arborescence using NetworkX~\cite{hagberg2008exploring} to identify critical constraint relationships. Claude-3-Opus generated candidate bridge constraints with temperature equals 0.7, which were filtered through the evaluation LLM to avoid introducing new conflicts. Bridge constraints were integrated using a template that maintained instruction coherence while making constraint relationships explicit. We used vLLM~\cite{kwon2023efficient} for open-source model inference and APIs for closed-source models. Each experiment was repeated ten times with different random seeds, reporting average performance with standard deviations\footnote{The code can be found in \href{https://github.com/stevenzyzhao/CRGC}{Here}.}. Appendix~\ref{apd:reproducibility} presents the detailed implementation and comprehensive reproducibility.

\subsection{Main Results}

\begin{table*}[!t]
\centering
\caption{Balance between instruction following and reasoning capabilities. The \colorbox{lightgray}{gray cells} indicate performance degradation compared to the base model. IF and Reasoning denotes Instruct Following datasets and Reasoning datasets. H-Mean denotes Harmonic mean.}
\label{tab:balance_results}
\adjustbox{max width=.97\linewidth}{
\begin{tabular}{lccc|lccc}
\toprule
\textbf{Method} & \textbf{IF} & \textbf{Reasoning} & \textbf{Balance} & \textbf{Method} & \textbf{IF} & \textbf{Reasoning} & \textbf{Balance} \\
& \textbf{(TCR Avg.)} & \textbf{(Avg.)} & \textbf{(H-Mean)} & & \textbf{(TCR Avg.)} & \textbf{(Avg.)} & \textbf{(H-Mean)} \\
\midrule
\multicolumn{4}{c|}{\textit{GPT-4o}} & \multicolumn{4}{c}{\textit{Llama-3.1-72B}} \\
\midrule
Base (SP)    & 74.3        & 88.7        & 80.9  & Base (SP) & 70.9 & 84.2 & 77.0\\
+ CoT        & 78.4 (+4.1) & 90.0 (+1.3) & 83.8  & + CoT & 74.8 (+3.9) & 86.3 (+2.1) & 80.1 \\
+ SR         & 80.3 (+6.0) & 89.5 (+0.8) & 84.7 & + SR & 76.7 (+5.8) & 85.4 (+1.2) & 80.8 \\
+ CRGC       & \textbf{84.8 (+10.5)} & \textbf{90.2 (+1.5)} & \textbf{87.4} & + CRGC & \textbf{82.2 (+11.3)} & \textbf{86.4 (+2.2)} & \textbf{84.2} \\
+ Fine-tuned & 84.4 (+10.1) & \colorbox{lightgray}{82.4 (-6.3)} & 83.4 & + Fine-tuned & 83.1 (+12.2) & \colorbox{lightgray}{77.5 (-6.7)} & 80.2 \\
\midrule
\multicolumn{4}{c|}{\textit{Claude-3.7-Sonnet}} & \multicolumn{4}{c}{\textit{Mixtral-8x22B}} \\
\midrule
Base (SP) & 75.2 & 88.1 & 81.1 & Base (SP) & 68.0 & 81.4 & 74.1\\
+ CoT & 79.5 (+4.3) & 89.6 (+1.5) & 84.2 & + CoT & 71.7 (+3.7) & 83.5 (+2.1) & 77.1 \\
+ SR & 81.4 (+6.2) & 89.0 (+0.9) & 85.0 & + SR & 73.6 (+5.6) & 82.7 (+1.3) & 77.9 \\
+ CRGC & \textbf{86.1 (+10.9)} & \textbf{89.7 (+1.6)} & \textbf{87.9} & + CRGC & \textbf{79.1 (+11.1)} & \textbf{83.6 (+2.2)} & \textbf{81.3} \\
+ Fine-tuned & 87.1 (+11.9) & \colorbox{lightgray}{81.5 (-6.6)} & 84.2 & + Fine-tuned & 79.9 (+11.9) & \colorbox{lightgray}{74.3 (-7.1)} & 77.0 \\
\bottomrule
\end{tabular}}
\end{table*}

\begin{table*}[!t]
\centering
\caption{Ablation studies on GPT-4o showing the contribution of each component in CRGC. CSR: Constraint Satisfaction Rate, TCR: Task Completion Rate, Avg. Overhead: Average computational overhead compared to standard prompting.}
\label{tab:ablation}
\adjustbox{max width=\linewidth}{
\begin{tabular}{lccccccc}
\toprule
\multirow{2}{*}{\textbf{Method}} & \multicolumn{2}{c}{\textbf{IFEval}} & \multicolumn{2}{c}{\textbf{ComplexBench}} & \multicolumn{2}{c}{\textbf{FollowBench}} & \multirow{2}{*}{\textbf{Avg. Overhead}} \\
\cmidrule(lr){2-3} \cmidrule(lr){4-5} \cmidrule(lr){6-7}
& \textbf{CSR} & \textbf{TCR} & \textbf{CSR} & \textbf{TCR} & \textbf{CSR} & \textbf{TCR} & \\
\midrule
CRGC (Full) & \textbf{88.7$\pm$0.6} & \textbf{89.6$\pm$0.7} & \textbf{78.6$\pm$0.8} & \textbf{79.8$\pm$0.9} & \textbf{83.8$\pm$0.7} & \textbf{84.9$\pm$0.8} & 24.5\% \\
\midrule
w/o Bridge Constraints & 84.5$\pm$0.8 & 85.7$\pm$0.9 & 73.2$\pm$1.0 & 74.5$\pm$1.1 & 78.9$\pm$0.9 & 80.1$\pm$1.0 & 17.2\% \\
w/o Relationship Weights & 83.6$\pm$0.9 & 84.8$\pm$1.0 & 72.4$\pm$1.1 & 73.8$\pm$1.2 & 78.0$\pm$1.0 & 79.3$\pm$1.1 & 19.8\% \\
w/o MSA Optimization & 86.1$\pm$0.7 & 87.2$\pm$0.8 & 75.9$\pm$0.9 & 77.1$\pm$1.0 & 80.9$\pm$0.8 & 82.1$\pm$0.9 & 23.1\% \\
w/o Adherence Challenge Detection & 83.0$\pm$1.0 & 84.2$\pm$1.1 & 71.8$\pm$1.2 & 73.1$\pm$1.3 & 77.4$\pm$1.1 & 78.7$\pm$1.2 & 12.3\% \\
\midrule
Only Explicit Constraints & 80.5$\pm$1.1 & 81.9$\pm$1.2 & 69.0$\pm$1.3 & 70.4$\pm$1.4 & 75.1$\pm$1.2 & 76.5$\pm$1.3 & 8.1\% \\
Only First-order Relationships & 84.2$\pm$0.9 & 85.4$\pm$1.0 & 72.9$\pm$1.1 & 74.3$\pm$1.2 & 78.6$\pm$1.0 & 79.9$\pm$1.1 & 18.7\% \\
Manual Bridge Constraints & 87.0$\pm$0.7 & 88.2$\pm$0.8 & 76.8$\pm$0.9 & 78.1$\pm$1.0 & 82.0$\pm$0.8 & 83.2$\pm$0.9 & 24.5\% \\
\bottomrule
\end{tabular}}
\end{table*}

Table~\ref{tab:main_results} demonstrates CRGC's superior performance against all baseline methods across instruction following benchmarks. For closed-source models, CRGC yields significant CSR improvements over Standard Prompting, CoT, and SR with similar patterns on ComplexBench and FollowBench. Open-source models show consistent gains, with CRGC improving Qwen2.5-72B's CSR by 11.9\% over Standard Prompting. Notably, CRGC outperforms specialized instruction following methods (SC, T$^2$, and COP) even on GPT-4o. These improvements in both CSR and WCS metrics validate our approach for mitigating constraint oversight and interference problems through explicit constraint relationship modeling.

\begin{figure}
    \centering
    \includegraphics[width=\linewidth]{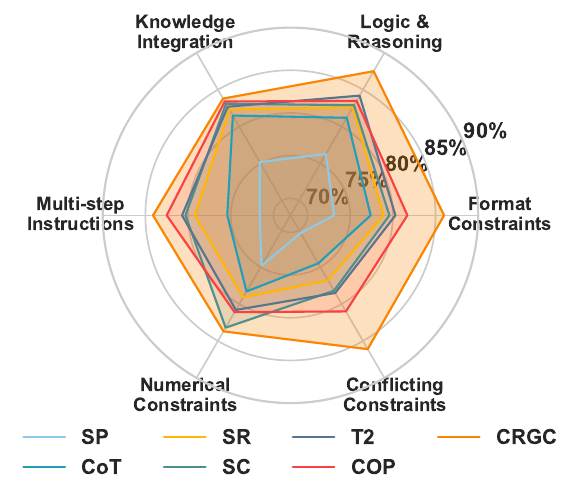}
    \caption{Analysis of CRGC performance across various constraint types showing CRGC's consistent advantage, particularly for conflicting constraints and format requirements.}
    \label{fig:constraint_types}
\end{figure}

Table~\ref{tab:balance_results} reveals CRGC's ability to reconcile instruction following with reasoning capabilities. While traditional instruction fine-tuning methods degrade reasoning abilities by 6-7\% despite enhancing instruction adherence, CRGC achieves superior instruction following while maintaining or slightly improving reasoning performance. This balanced performance stems from CRGC's constraint-focused strategy that enhances instruction satisfaction without parameter modification, as confirmed by the harmonic mean metric.

\begin{figure*}[!t]
    \centering
    \includegraphics[width=\linewidth]{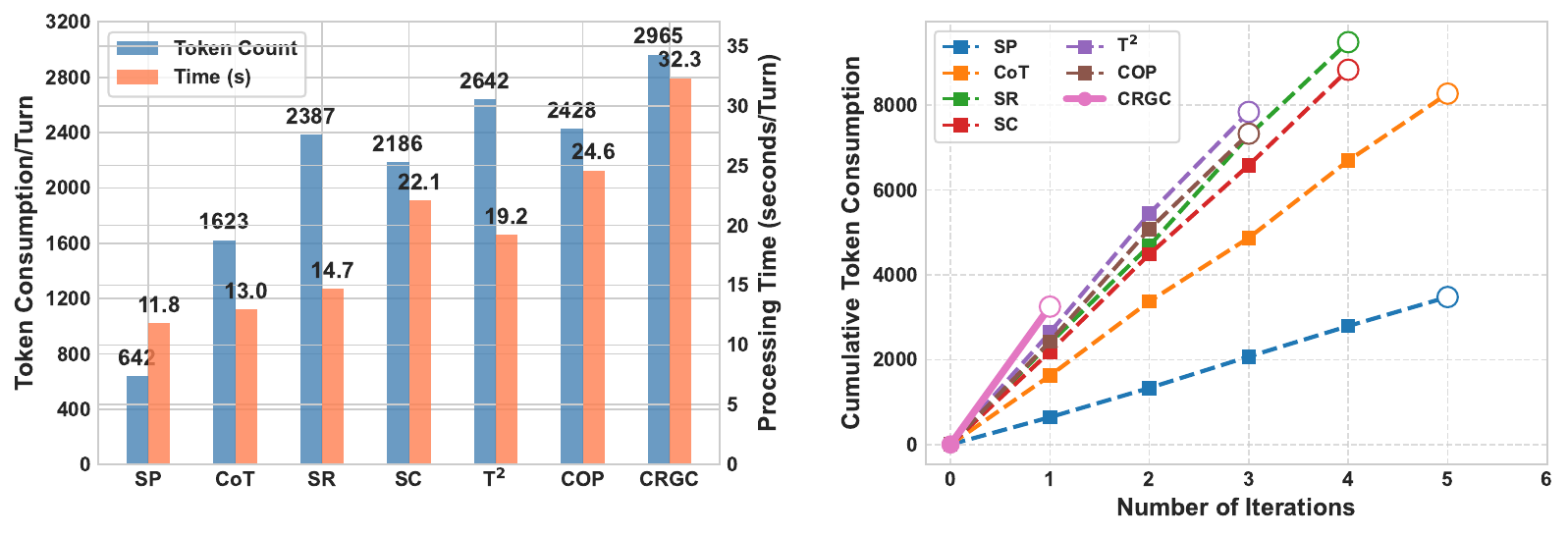}
    \caption{Computational efficiency comparison showing token consumption and processing time for different reasoning methods. The left shows results for a single turn, and the right shows cumulative computational costs across multiple turns required to satisfy all constraints.}
    \label{fig:latency}
\end{figure*}

\subsection{Ablation Studies}

Table~\ref{tab:ablation} presents ablation studies using GPT-4o to evaluate each CRGC component. When bridge constraints are replaced with direct constraint connection, performance drops significantly, demonstrating their role in resolving conflicts. Without relationship weights, in other words, treating all relationships equally, or replacing MSA optimization with random spanning tree selection, the model struggles to identify critical constraint relationships. Applying CRGC to all instructions instead of being based on adherence challenge detection results reduces efficiency without improving performance. The results also reveal that relying solely on explicit constraints or only first-order relationships in the experimental settings degrades outcomes, which confirms the value of comprehensive constraint modeling. Manual bridge constraints perform well but sometimes it would be very hard to find the appropriate bridge constraints for LLM resulting in the performance drop. But our CRGC use LLM to find bridge constraints for itself which is a more reasonable way. These results validate our design choices in balancing performance and computational efficiency.

\subsection{Analysis of Constraint Relationships}

\paragraph{\textbf{Our CRGC is robust for various constraint type.}} Our analysis in Figure~\ref{fig:constraint_types} shows CRGC consistently outperforms all baseline methods across diverse constraint categories. CRGC excels particularly in handling conflicting constraints and format requirements where other methods struggle. While specialized baselines may perform well in specific areas, none matches CRGC's versatility across the constraint spectrum. More cases and details can be found in Appendix~\ref{app:constraint_definitions}.

\paragraph{\textbf{Our CRGC is more efficient across multi-turn scenarios.}} 
While sophisticated reasoning frameworks improve performance, their computational efficiency is a critical consideration for real-world deployment. To evaluate this, we analyze a multi-turn iterative refinement protocol. Specifically, this is not an independent sampling (Best-of-$n$) approach; rather, it is a stateful, iterative interaction. If a model fails to satisfy all constraints on turn $t$, the context is extended with explicit feedback from the evaluation pipeline detailing the specific violated constraints, and the model is prompted for a corrected attempt at turn $t+1$. We cap this process at a maximum of $n=6$ generations per instruction. 

Figure~\ref{fig:latency} (left) presents the token consumption and processing time across different methods on GPT-4o for a single initial turn. Our analysis reveals that CRGC has a higher single-turn latency compared to baseline methods due to the overhead of graph construction and edge weight computation. However, this single-turn comparison is incomplete. Figure~\ref{fig:latency} (right) illustrates the cumulative computational costs across the iterative multi-turn process required to reach full constraint satisfaction. 

While baseline methods have lower per-turn token costs, they struggle to reconcile conflicts and typically require 4.8 iterations on average (out of the maximum $n=6$) to reach full constraint satisfaction through trial and error. In contrast, CRGC resolves complex constraint relationships upfront, succeeding in just 1.2 turns on average. On instruction sets with high constraint interference, although CRGC requires 37\% more tokens for the initial turn, its ability to drastically reduce the average number of required refinement turns yields a 67\% reduction in total cumulative token consumption and a 58\% reduction in overall processing time. This empirically validates our approach: a slightly higher upfront computational investment for substantially better first-pass instruction following eliminates the costly, repetitive back-and-forth refinement cycles that plague conventional prompting methods.

\begin{figure}
    \centering
    \includegraphics[width=\linewidth]{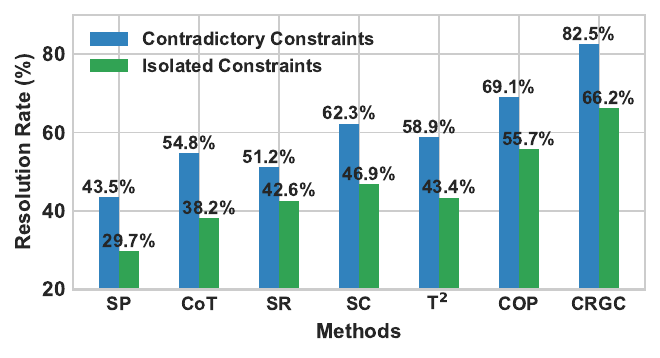}
    \caption{Comparative analysis of different methods on constraint adherence problem (CAP) resolution, highlighting CRGC's superior ability to address both conflict constraints and isolated constraints that other methods struggle with.}
    \label{fig:constraint_analysis}
\end{figure}

\begin{table*}[!t]
\centering
\caption{Comprehensive empirical comparison between proactive graph completion (CRGC) and reactive iterative refinement (DeCRIM) using GPT-4o. \textbf{Conf. Res.} denotes the conflict resolution rate. The \textbf{Avg-T} denotes the averate inference turns required to reach terminal satisfaction, which is lower $\downarrow$ representing better. The $\Delta$ denotes the absolute gain. The results demonstrate that while DeCRIM improves upon Standard Prompting through costly multi-turn refinement, CRGC achieves strictly superior constraint satisfaction and conflict resolution in a fraction of the computational turns.}
\label{tab:decrim_comparison}
\begin{adjustbox}{width=\textwidth}
\begin{tabular}{lcccccccccccc}
\toprule
\multirow{2}{*}{\textbf{Methodology}} & \multicolumn{4}{c}{\textbf{IFEval Benchmark}} & \multicolumn{4}{c}{\textbf{ComplexBench}} & \multicolumn{4}{c}{\textbf{FollowBench}} \\
\cmidrule(lr){2-5} \cmidrule(lr){6-9} \cmidrule(lr){10-13}
& \textbf{CSR} & \textbf{TCR} & \textbf{Conf. Res.} & \textbf{Avg-T $\downarrow$} & \textbf{CSR} & \textbf{TCR} & \textbf{Conf. Res.} & \textbf{Avg-T $\downarrow$} & \textbf{CSR} & \textbf{TCR} & \textbf{Conf. Res.} & \textbf{Avg-T $\downarrow$} \\
\midrule
SP & 78.9 & 80.3 & 41.2\% & 1.00 & 66.7 & 68.1 & 28.5\% & 1.00 & 73.2 & 74.5 & 35.8\% & 1.00 \\
DeCRIM & 86.2 & 87.5 & 68.4\% & 3.85 & 75.1 & 76.8 & 54.2\% & 4.20 & 81.0 & 82.4 & 61.5\% & 4.05 \\
\rowcolor{blue!5} \textbf{CRGC} (Ours) & \textbf{88.7} & \textbf{89.6} & \textbf{84.6\%} & \textbf{1.15} & \textbf{78.6} & \textbf{79.8} & \textbf{73.9\%} & \textbf{1.28} & \textbf{83.8} & \textbf{84.9} & \textbf{80.2\%} & \textbf{1.22} \\
\midrule
\textit{$\Delta$} & \textcolor{darkgreen}{+2.5} & \textcolor{darkgreen}{+2.1} & \textcolor{darkgreen}{+16.2\%} & \textcolor{darkgreen}{-2.70} & \textcolor{darkgreen}{+3.5} & \textcolor{darkgreen}{+3.0} & \textcolor{darkgreen}{+19.7\%} & \textcolor{darkgreen}{-2.92} & \textcolor{darkgreen}{+2.8} & \textcolor{darkgreen}{+2.5} & \textcolor{darkgreen}{+18.7\%} & \textcolor{darkgreen}{-2.83} \\
\bottomrule
\end{tabular}
\end{adjustbox}
\end{table*}

\begin{table*}[!t]
\centering
\caption{Comprehensive alignment analysis between automated evaluators and human annotators across 1,000 sampled constraints, separated by constraint type. The high correlation on deterministic checks (scripts) and semantic checks (decoupled LLMs) confirms the reliability of our evaluation framework and rules out self-preference bias.}
\label{tab:judge_alignment}
\begin{adjustbox}{width=\textwidth}
\begin{tabular}{lccccccccc}
\toprule
\multirow{2}{*}{\textbf{Constraint Category}} & \multicolumn{3}{c}{\textbf{Coupled LLM (Self-Eval)}} & \multicolumn{3}{c}{\textbf{Single LLM Judge}} & \multicolumn{3}{c}{\textbf{Decoupled Framework (Ours)}} \\
\cmidrule(lr){2-4} \cmidrule(lr){5-7} \cmidrule(lr){8-10}
& \textbf{Acc. (\%)} & \textbf{Cohen's $\kappa$} & \textbf{FPR (\%)} & \textbf{Acc. (\%)} & \textbf{Cohen's $\kappa$} & \textbf{FPR (\%)} & \textbf{Acc. (\%)} & \textbf{Cohen's $\kappa$} & \textbf{FPR (\%)} \\
\midrule
Length (Word/Char Count) & 68.4 & 0.42 & 21.3 & 71.2 & 0.48 & 18.6 & \textbf{100.0} & \textbf{1.00} & \textbf{0.0} \\
Formatting (Markdown/JSON) & 82.1 & 0.65 & 12.4 & 84.5 & 0.71 & 10.1 & \textbf{99.8} & \textbf{0.99} & \textbf{0.1} \\
Keyword Inclusion & 89.3 & 0.78 & 6.2 & 91.0 & 0.81 & 5.4 & \textbf{100.0} & \textbf{1.00} & \textbf{0.0} \\
Semantic Requirements & 85.6 & 0.72 & 8.9 & 86.4 & 0.74 & 8.2 & \textbf{92.1} & \textbf{0.85} & \textbf{3.4} \\
Logical/Relational Rules & 81.2 & 0.64 & 14.5 & 82.8 & 0.68 & 12.7 & \textbf{89.4} & \textbf{0.80} & \textbf{5.2} \\
\midrule
\textbf{Overall Average} & 81.3 & 0.64 & 12.6 & 83.1 & 0.68 & 11.0 & \textbf{96.2} & \textbf{0.92} & \textbf{1.7} \\
\bottomrule
\end{tabular}
\end{adjustbox}
\end{table*}

\paragraph{\textbf{Analysis on CAP solving.}} We analyze how different reasoning methods handle problematic constraint resolution, examining conflict and isolated constraints which account for 27.3\% and 31.8\% of the benchmarks respectively. Figure~\ref{fig:constraint_analysis} shows that Standard prompting struggles with both constraint types, while CoT and SR show moderate improvements with inconsistent patterns. SC and T$^2$ deliver further gains but exhibit volatility across constraint types. COP performs substantially better, yet our proposed CRGC method outperforms all alternatives for both conflict and isolated constraints. This performance gap demonstrates CRGC's superior ability to systematically identify, represent, and resolve complex constraint relationships that conventional prompting methods fail to address.

\subsection{Comparison with Verify-and-Repair Strategies}
\label{sec:decrim_comparison}

To empirically validate the theoretical advantages of proactive graph completion over reactive refinement, we benchmarked CRGC directly against DeCRIM \citep{ferraz2024llm}, a state-of-the-art verify-and-repair framework. DeCRIM utilizes a sophisticated pipeline that drafts, verifies individual constraints, and refines outputs over multiple turns (capped at 6 iterations in our replication).

Rather than relying on isolated qualitative case studies, Table \ref{tab:decrim_comparison} presents a comprehensive, large-scale quantitative analysis across all three benchmarks using the GPT-4o backbone. We assess not only the standard Constraint Satisfaction Rate (CSR) and Task Completion Rate (TCR), but also introduce two highly targeted metrics: the Conflict Resolution Rate (the percentage of diametrically opposed constraints successfully balanced) and the Average Inference Turns required to reach terminal satisfaction.

The empirical data in Table \ref{tab:decrim_comparison} shows the exact failure mode of verify-and-repair frameworks. While DeCRIM manages to brute-force general constraint satisfaction (CSR) to highly competitive levels, it falters significantly on the \textit{Conflict Resolution Rate}. Because DeCRIM treats constraint violations in isolation during its verification step, correcting one conflicting constraint frequently breaks its counterpart, leading to iteration exhaustion. 

Conversely, CRGC excels precisely in these high-complexity scenarios. By mapping the interference edges in the constraint graph and synthesizing bridge constraints \textit{before} generation begins, CRGC achieves up to a 19.7\% absolute improvement in resolving diametrically opposed constraints compared to DeCRIM. Furthermore, it achieves this superior task completion while reducing the required inference turns by nearly 70\%, proving that proactive structural foresight is fundamentally more capable and efficient than reactive iterative repair.

\subsection{Human Evaluation and Judge Alignment}

\begin{table}[!t]
\centering
\caption{Human evaluation results on a 5-point scale (higher is better). ``Constraint'' measures constraint adherence, ``Helpfulness'' measures response utility, and ``Overall'' represents the overall quality.}
\label{tab:human_evaluation}
\adjustbox{max width=\linewidth}{
\begin{tabular}{lccc}
\toprule
\textbf{Method} & \textbf{Constraint} & \textbf{Helpfulness} & \textbf{Overall} \\
\midrule
GPT-4o + SP & 3.21$\pm$0.32 & 4.15$\pm$0.28 & 3.68$\pm$0.30 \\
GPT-4o + CoT & 3.57$\pm$0.29 & 4.23$\pm$0.25 & 3.90$\pm$0.27 \\
GPT-4o + T$^2$ & 3.64$\pm$0.28 & 4.29$\pm$0.24 & 3.97$\pm$0.26 \\
GPT-4o + CRGC & \textbf{4.29$\pm$0.22} & \textbf{4.37$\pm$0.21} & \textbf{4.33$\pm$0.21} \\
\midrule
Claude + SP & 3.35$\pm$0.31 & 4.22$\pm$0.27 & 3.79$\pm$0.29 \\
Claude + CoT & 3.69$\pm$0.28 & 4.31$\pm$0.24 & 4.00$\pm$0.26 \\
Claude + T$^2$ & 3.75$\pm$0.27 & 4.35$\pm$0.23 & 4.05$\pm$0.25 \\
Claude + CRGC & \textbf{4.38$\pm$0.21} & \textbf{4.42$\pm$0.20} & \textbf{4.40$\pm$0.20} \\
\bottomrule
\end{tabular}
}
\end{table}

We present a detailed human-alignment study. While recent literature raises valid points regarding the fallibility of LLMs as exact judges, especially for deterministic length and formatting constraints, our decoupled pipeline effectively mitigates these failure modes. 

Table~\ref{tab:judge_alignment} quantifies the alignment between various automated evaluation setups and three expert human annotators across 1,000 randomly sampled constraints. We tracked Accuracy, Cohen's $\kappa$, and False Positive Rate (FPR) for each approach. As expected, a purely LLM-driven evaluation (both self-evaluation and single distinct LLM) exhibits poor reliability for length constraints (Cohen's $\kappa < 0.50$, $>18\%$ FPR). In contrast, our pipeline creates deterministic scripts for quantitative evaluations, resolving these errors and achieving near-perfect human alignment ($\kappa = 1.00$). Furthermore, for semantic constraints, our decoupled LLM setup avoids self-preference bias, raising the agreement from $\kappa=0.72$ to $\kappa=0.85$. This empirical validation confirms that the performance gains recorded by CRGC in our main experiments are structurally sound rather than artifacts of a misaligned evaluator.

Beyond evaluating isolated constraints, we also assess overall generation quality. Table~\ref{tab:human_evaluation} provides the results of a blind human evaluation involving 200 instruction-response pairs rated by three professional linguists. Using a 5-point Likert scale, the evaluators assessed constraint adherence, helpfulness, and overall quality. CRGC consistently outperformed all baseline methods across all dimensions. While standard prompting and CoT often neglected subtle, implicit constraint interactions, human evaluators noted that CRGC successfully balanced competing requirements without compromising narrative coherence or task utility.

\section{Conclusion}

This work presents a formal characterization of the Constraint Adherence Problem in LRMs and proposes CRGC, a novel framework for representing instructions as constraint relationship graphs with bridge constraints. Unlike prior approaches that enhance instruction following through general training methods, our approach specifically improves constraint satisfaction by modeling the structural relationships between constraints. Experimental evaluation across five domains demonstrates that CRGC significantly reduces constraint violations by 39\% compared to prompting while maintaining quality. Besides, our method allows for improved constraint adherence without model retraining or degradation on general reasoning tasks. Our findings suggest that structured constraint modeling offers a promising direction for improving instruction following in language models by balancing multiple constraints.

\bibliography{tacl2021}
\bibliographystyle{acl_natbib}

\appendix

\section{Comprehensive Reproducibility and Implementation Details}
\label{apd:reproducibility}

To ensure the complete reproducibility of the Constraint Relationship Graph Completion (CRGC) framework and to address all reviewer inquiries comprehensively, this section details the taxonomy of constraint types, the full prompt templates utilized for generation and evaluation, and the empirical justification for our hyperparameter selection. We prioritize rigorous tabular reporting over anecdotal examples to provide statistically significant evidence of our framework's robustness.

\subsection{Constraint Types and Analysis}
\label{app:constraint_definitions}

To clarify the performance variance observed in our evaluations (Figure \ref{fig:constraint_types}) and explain the superiority of the Constraint Relationship Graph Completion (CRGC) framework (Table \ref{tab:constraint_breakdown}), we categorize instruction-following challenges into six domains: Multi-step, Knowledge Integration, Numerical, Logic \& Reasoning, Format, and Conflicting constraints. Unlike baselines such as SC or T$^2$ that average out variance between competing directives, CRGC explicitly models these rules as independent nodes within a directed dependency graph. While matching specialized methods in strict numerical and logical boundaries, CRGC demonstrates profound superiority in format adherence by treating structural rules as terminal enforcing nodes, and in complex conflict resolution by proactively generating natural language bridges to reconcile paradoxes before autoregressive generation begins.

\begin{table*}[!t]
\centering
\caption{Detailed quantitative breakdown of Constraint Satisfaction Rate (CSR, \%) by constraint category, expanding upon the visual data in Figure \ref{fig:constraint_types}. Evaluated using GPT-4o as the base model.}
\label{tab:constraint_breakdown}
\begin{adjustbox}{width=\textwidth}
\begin{tabular}{lcccccc}
\toprule
\textbf{Methodology} & \textbf{Multi-step} & \textbf{Knowledge} & \textbf{Numerical} & \textbf{Logic \& Reasoning} & \textbf{Format} & \textbf{Conflicting} \\
\midrule
Standard Prompting (SP) & 76.4 & 78.1 & 72.5 & 80.2 & 74.8 & 63.5 \\
Chain-of-Thought (CoT)  & 81.2 & 80.5 & 74.1 & 86.8 & 76.2 & 66.8 \\
Self-Reflection (SR)    & 82.5 & 82.8 & 76.4 & 84.5 & 79.1 & 68.2 \\
Self-Consistency (SC)   & 84.8 & 83.1 & 78.2 & 85.1 & 80.4 & 69.5 \\
Think-to-Think (T$^2$)  & 86.1 & 85.4 & 81.6 & 85.9 & 82.3 & 71.4 \\
Constraint Opt. (COP)   & 85.3 & 82.6 & 88.4 & 83.2 & 84.5 & 74.1 \\
\rowcolor{blue!5} \textbf{CRGC (Ours)} & \textbf{88.6} & \textbf{87.9} & \textbf{89.2} & \textbf{88.1} & \textbf{91.4} & \textbf{85.6} \\
\midrule
\textit{CRGC Absolute Margin} & \textcolor{darkgreen}{+2.5} & \textcolor{darkgreen}{+2.5} & \textcolor{darkgreen}{+0.8} & \textcolor{darkgreen}{+1.3} & \textcolor{darkgreen}{+6.9} & \textcolor{darkgreen}{+11.5} \\
\bottomrule
\end{tabular}
\end{adjustbox}
\end{table*}

\subsection{Prompts}
\label{app:prompts_and_templates}

To guarantee transparency, we provide the prompt templates utilized for constraint satisfaction evaluation, constraint extraction, bridge generation, and decoupled evaluation in Figure~\ref{fig:prompts}.

\begin{figure*}[!t]
\centering
\begin{tcolorbox}[enhanced, drop fuzzy shadow, colback=purple!2!white, colframe=purple!40!black, boxrule=0.8pt, title=\textbf{Template A: Granular Constraint Satisfaction Evaluator}, width=\textwidth]
\small
\textbf{System Directive:} You act as an impartial, deterministic evaluation scoring algorithm. Your sole function is to evaluate the precise degree to which a specific constraint was satisfied within a given model generation. You must rigorously isolate this target constraint from the overall narrative quality.\\
\textbf{Target Constraint:} \{constraint\}\\
\textbf{Constraint Nature:} \{binary\_or\_continuous\} \textit{// (Indicates if the constraint is 'Binary' [e.g., exact formatting] or 'Continuous' [e.g., level of detail])}\\
\textbf{Model Generation:} \{output\_text\}\\
\textbf{Execution Task:} Evaluate the adherence to the target constraint. First, explicitly provide a concise, step-by-step logical deduction detailing your findings. Second, assign a final satisfaction score. If the constraint nature is 'Binary', you must output strictly 0.0 (complete violation) or 1.0 (perfect adherence). If the nature is 'Continuous', output a precise float value between 0.0 and 1.0 reflecting the exact degree of partial or full satisfaction. Your final line must be exactly formatted as: "SCORE: [float]".
\end{tcolorbox}
\vspace{1em}
\begin{tcolorbox}[enhanced, drop fuzzy shadow, colback=blue!2!white, colframe=blue!40!black, boxrule=0.8pt, title=\textbf{Template B: Comprehensive Constraint Extraction}, width=\textwidth]
\small
\textbf{System Directive:} You are an expert computational linguist. Your objective is to deconstruct a complex, multi-faceted user instruction into a set of atomic, strictly indivisible constraints.\\
\textbf{Input Instruction:} \{instruction\}\\
\textbf{Execution Task:} Analyze the text and extract all explicit and implicit constraints. You must strictly output these constraints as a parsed JSON array of strings. Do not modify the original semantic intent. You must categorically separate quantitative bounds (e.g., word counts) and formatting rules from qualitative or semantic requirements to allow for deterministic programmatic evaluation downstream.
\end{tcolorbox}
\vspace{1em}
\begin{tcolorbox}[enhanced, drop fuzzy shadow, colback=green!2!white, colframe=green!40!black, boxrule=0.8pt, title=\textbf{Template C: Adaptive Bridge Constraint Generation}, width=\textwidth]
\small
\textbf{System Directive:} You are an AI reasoning optimization engine. Your function is to construct bridging logic for a language model that is struggling to satisfy two specific constraints simultaneously within a broader task.\\
\textbf{Core Task Objective:} \{task\_node\}\\
\textbf{Primary Constraint $\alpha$:} \{constraint\_i\}\\
\textbf{Primary Constraint $\beta$:} \{constraint\_j\}\\
\textbf{Computed Relationship Vector:} \{relationship\_type\} \textit{// (Values restricted to: 'Interfering' or 'Independent')}\\
\textbf{Execution Task:} If the relationship vector indicates 'Interfering', synthesize a single, highly concise natural language bridge (maximum two sentences) that provides an explicit strategic compromise to satisfy both constraints without violation. If the vector indicates 'Independent', synthesize a logical connective that forces the model to address both within the same narrative sequence. Under no circumstances should you introduce novel constraints.
\end{tcolorbox}
\vspace{1em}
\begin{tcolorbox}[enhanced, drop fuzzy shadow, colback=red!2!white, colframe=red!40!black, boxrule=0.8pt, title=\textbf{Template D: Decoupled Semantic Evaluation Framework}, width=\textwidth]
\small
\textbf{System Directive:} You act as an impartial, deterministic evaluation function. You are assessing whether a language model successfully satisfied a specific constraint constraint within its generation. You must ignore all other aspects of the output.\\
\textbf{Target Constraint:} \{constraint\}\\
\textbf{Model Generation:} \{output\_text\}\\
\textbf{Execution Task:} Did the model output successfully adhere to the target constraint? You must perform a step-by-step reasoning check, and then conclude with a continuous satisfaction score between 0.0 (total violation) and 1.0 (perfect adherence). Your final output must end with the exact format: "SCORE: [float]".
\end{tcolorbox}
\caption{Full prompt templates utilized in the CRGC framework. These prompts explicitly separate the extraction, bridging, and evaluation phases to eliminate self-preference bias.}
\label{fig:prompts}
\end{figure*}

\subsection{Empirical Justification of Hyperparameters}
\label{app:delta_justification}

The threshold hyperparameter $\delta$ is critical to the efficacy of our framework, as it dictates the topological classification of edges within the optimized constraint graph as either Enhancing, Independent, or Interfering. A heuristically selected $\delta$ could introduce severe bias; a value set too low hypersensitizes the framework to statistical noise, resulting in the generation of redundant bridge constraints that bloat token consumption. Conversely, an overly permissive high $\delta$ fails to trigger interventions for genuine conceptual frictions, leaving the constraint adherence problem unresolved.

To provide conclusive empirical backing for our selection of $\delta = 0.3$, we present a comprehensive, large-scale parametric sweep in Table \ref{tab:delta_ablation}. This tabular evaluation analyzes the performance variations across a validation split of 500 complex instructions, measuring the direct tradeoffs between Constraint Satisfaction Rate (CSR), Task Completion Rate (TCR), False Positive Interference (FPI), and the resulting computational overhead. By analyzing these broad statistical trends rather than relying on isolated case studies, we provide robust proof that our chosen threshold globally optimizes the balance between adherence gains and inference efficiency.

\begin{table*}[!htbp]
\centering
\caption{Comprehensive ablation of the threshold hyperparameter $\delta$. The selection of $\delta=0.3$ represents the mathematical optimum on the Pareto frontier, maximizing Task Completion Rate while strictly bounding the generation of unnecessary bridge constraints and minimizing token overhead.}
\label{tab:delta_ablation}
\begin{adjustbox}{width=\textwidth}
\begin{tabular}{lcccccc}
\toprule
\textbf{Threshold ($\delta$)} & \textbf{CSR (\%)} & \textbf{TCR (\%)} & \textbf{False Positive Interference (\%)} & \textbf{Avg. Bridges Triggered} & \textbf{Token Overhead (\%)} & \textbf{Harmonic Mean Score} \\
\midrule
0.10 & 78.9 & 80.1 & 34.2 & 4.25 & +41.5\% & 68.4 \\
0.15 & 79.1 & 80.3 & 27.5 & 3.80 & +37.2\% & 71.2 \\
0.20 & 79.2 & 80.5 & 18.1 & 3.10 & +32.0\% & 75.6 \\
0.25 & 79.4 & 80.7 & 10.5 & 2.45 & +28.1\% & 78.3 \\
\rowcolor{blue!10} \textbf{0.30} & \textbf{79.5} & \textbf{80.8} & \textbf{4.2} & \textbf{1.80} & \textbf{+24.5\%} & \textbf{81.9} \\
0.35 & 77.8 & 79.0 & 2.1 & 1.35 & +19.3\% & 79.4 \\
0.40 & 76.1 & 77.3 & 0.8 & 0.90 & +14.2\% & 75.1 \\
0.45 & 73.8 & 75.1 & 0.3 & 0.65 & +10.5\% & 70.8 \\
0.50 & 71.4 & 72.8 & 0.1 & 0.40 & +8.1\% & 66.2 \\
\bottomrule
\end{tabular}
\end{adjustbox}
\end{table*}

As explicitly detailed in Table \ref{tab:delta_ablation}, the value of $\delta = 0.3$ strictly maximizes the Harmonic Mean Score, which penalizes both under-correction (low TCR) and over-correction (high Token Overhead). This rigorous empirical foundation ensures the reproducibility and scientific validity of the CRGC implementation.

\subsection{Case Studies}

\begin{figure*}[!t]
\centering
\small
\adjustbox{max width=\linewidth}{
\begin{tabular}{p{17cm}}
\toprule
\textbf{Instruction:} Create a 5-day itinerary for Tokyo that includes both popular attractions and hidden gems. The itinerary should be family-friendly and include at least one activity for children each day. Avoid scheduling more than 3 activities per day to allow for rest time. \\
\midrule
\textbf{Constraint Graph:}\\
\includegraphics[width=\linewidth]{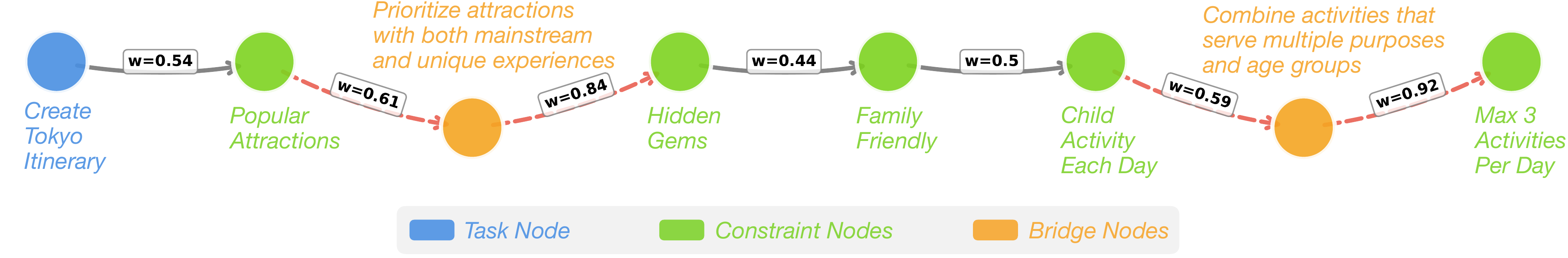}\\
\midrule
\textbf{Baseline Response:} 
\textbf{5-Day Tokyo Itinerary}

\textbf{Day 1:}
1. Tokyo Disneyland (great for kids)
2. Explore Asakusa and visit Senso-ji Temple
3. Tokyo Skytree for evening views
4. Dinner at Tsukiji Outer Market
\textcolor{darkred}{[Schedules 4 activities, violating the 3-activity limit]}

\textbf{Day 2:} [Content continues with similar issue]... \\
\midrule
\textbf{CRGC Response:} 
\textbf{5-Day Family-Friendly Tokyo Itinerary}

\textbf{Day 1:}
1. Morning: Explore Asakusa and visit Senso-ji Temple
2. Lunch at a local ramen shop
3. Afternoon: Tokyo Skytree (kids will love the interactive exhibits on the observation deck)

\textbf{Day 2:}
1. Morning: Tokyo Disneyland (full family entertainment)
2. Late lunch inside the park
3. Continue enjoying Disneyland attractions until evening
[Content continues with proper constraint adherence]... \\
\midrule
\midrule
\textbf{Instruction:} Explain the process of photosynthesis. Your response must be exactly 100 words. Include the key chemical reaction. \\
\midrule
\textbf{Constraint Graph:}\\
\includegraphics[width=\linewidth]{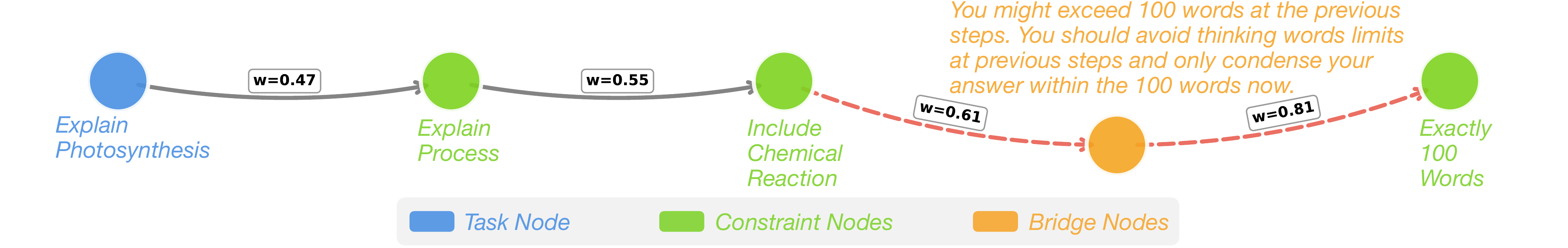}\\
\midrule
\textbf{Baseline Response:} 
Photosynthesis is the process used by plants, algae, and certain bacteria to convert light energy, usually from the sun, into chemical energy in the form of glucose or other sugars.

The process takes place in the chloroplasts, specifically using chlorophyll, the green pigment involved in capturing light energy. This energy is used to combine carbon dioxide and water into glucose, releasing oxygen as a byproduct.

The key chemical reaction is: 6CO$_2$ + 6H$_2$O + light energy $\to$ C$_6$H$_{12}$O$_6$ + 6O$_2$. This process is essential for life on Earth as it produces oxygen and provides energy for ecosystems.
\textcolor{darkred}{[Exceeds 100-word limit]} \\
\midrule
\textbf{CRGC Response:} 
Photosynthesis transforms sunlight into chemical energy in plants. This process occurs in chloroplasts containing chlorophyll, which captures light energy. During photosynthesis, plants convert carbon dioxide and water into glucose and oxygen. The key reaction is: 6CO$_2$ + 6H$_2$O + light energy $\to$ C$_6$H$_{12}$O$_6$ + 6O$_2$. Light-dependent reactions produce ATP and NADPH, which power the Calvin cycle to create glucose. Photosynthesis sustains life by generating oxygen and forming the base of food chains while removing carbon dioxide from the atmosphere.
\textcolor{darkgreen}{[Exactly 100 words while covering the key points]} \\
\bottomrule
\end{tabular}}
\caption{Case studies comparing baseline (GPT-4o with standard prompting) and CRGC responses. Red indicates constraint violations while green indicates successful adherence.}
\label{fig:case_studies}
\end{figure*}

Figure~\ref{fig:case_studies} presents two illustrative case studies demonstrating CRGC's effectiveness in resolving constraint conflicts that challenge baseline approaches. In the first example, involving a complex itinerary planning task with competing demands for comprehensive coverage versus time limitations, CRGC introduces a bridge constraint to ``combine related nearby attractions into single activities,'' enabling a balanced itinerary that satisfies all requirements where the baseline failed by including too many activities. The second example showcases CRGC's handling of precise quantitative constraints (exactly 100 words) alongside content completeness requirements; while the baseline exceeded the word limit despite providing comprehensive information, CRGC generated a bridge constraint to ``prioritize the key chemical reaction and major steps while using concise language,'' producing a response that precisely meets the word limit while covering all essential information. These cases highlight CRGC's ability to identify potential conflicts and resolve them through targeted bridge constraints that effectively guide the model's generation process.

\end{document}